\documentclass[mlabstract,onecolumn]{jmlr}

\usepackage{booktabs}
\usepackage{float}
\hypersetup{hypertexnames=false}

\jmlrvolume{}
\firstpageno{1}
\jmlryear{2026}
\jmlrworkshop{Symmetry and Geometry in Neural Representations}
\jmlrproceedings{PMLR}{Extended Abstract Track}

\DeclareMathOperator{\StopGrad}{StopGrad}
\newcommand{\R}{\mathbb{R}}
\newcommand{\E}{\mathbb{E}}
\newcommand{\calL}{\mathcal{L}}
\newcommand{\norm}[1]{\lVert #1\rVert}
\newcommand{\SD}{W_1}

\title[Euclidean Drifting Models]{Euclidean Drifting is a Strong Topological Baseline}

\author[Thompson et al.]{%
 \Name{Luke Thompson} \quad \Name{Xiaoyu Li} \quad \Name{Dai Shi} \quad \Name{Andi Han} \quad \Name{Junbin Gao}\\
 \addr The University of Sydney \quad Correspondence: \texttt{luke.thompson@sydney.edu.au}
}

\begin{document}

\raggedbottom

\maketitle

\begin{abstract}
Topological flow matching and Schr\"odinger bridge models build the Hodge Laplacian into continuous-time dynamics for signals on graphs and simplicial complexes. We ask whether it helps Drifting Models, which generate a sample in one network evaluation. We add topology in two separable places: the loss, through a Hodge--Sobolev metric with heat-kernel smoothing, and the predictor, through graph propagation. With affinities fixed, the Hodge loss only rescales each spectral component of the Euclidean update by a positive factor. In seven matched comparisons on five benchmarks, the Hodge loss never improves significantly, and sweeping its two hyperparameters finds no setting that helps consistently. Replacing the predictor's Laplacian with a randomly permuted copy does not change $W_1$. On brain data, Euclidean Drifting with a graph predictor beats our I-TFM reproduction with one network evaluation instead of $1{,}000$.
\end{abstract}

{\leftskip 20pt\rightskip 20pt\small\noindent\textbf{Code:} \url{https://github.com/luke-a-thompson/Topological-Drifting}.\par}


\section{Introduction}
\label{sec:introduction}

Many scientific signals live on the nodes or edges of a fixed graph or simplicial complex, and the Hodge Laplacian $L$ gives them a notion of frequency and a heat flow \citep{schaub2021higher}. Topological flow matching (TFM) and topological Schr\"odinger bridge matching (TSBM) add the drift $-\kappa L X_t$ to a continuous-time reference process, which shapes both the learned paths and the sampling trajectories at the cost of hundreds of network evaluations per sample \citep{yang2025topological,wyrwal2026topological}.

Drifting Models generate a sample with one network evaluation \citep{deng2026drifting}. They have no path for a heat drift to act on, so it is unclear whether topology helps them. Because the loss and predictor can be changed independently, we compare Euclidean and Hodge losses with the architecture fixed, and four graph predictors with the loss fixed.

\paragraph{Contributions.} (i) The fixed-affinity Hodge loss is a positive spectral preconditioner (Proposition~\ref{prop:preconditioner}). (ii) It gives no significant improvement in seven comparisons, and no setting in a hyperparameter sweep helps across tasks. (iii) Permuting the predictor's Laplacian does not change performance. (iv) With a GCN predictor, Euclidean Drifting beats our I-TFM reproduction on brain data.

\section{Separating topology in the loss and predictor}
\label{sec:method}

Let $x\in\R^d$ be a cochain of fixed degree on a finite graph or simplicial complex $K$, and let $L=L^\top\succeq0$ be its Hodge Laplacian in coordinates where it is self-adjoint (Table~\ref{tab:tasks}). A generator $f_\theta$ pushes a source law $\mu_0$ forward to $q=(f_\theta)_\#\mu_0$, which should match a target law $p$. Euclidean Drifting forms the population field
\begin{equation}
V_E(z)=\E_{y\sim p}\!\left[a_{p,q}(z,y)(y-z)\right]-\E_{y\sim q}\!\left[b_{p,q}(z,y)(y-z)\right],
\label{eq:euclidean-field}
\end{equation}
where $a_{p,q}$ and $b_{p,q}$ are balanced radial-kernel affinities with $a_{p,q}(z,y)=b_{q,p}(z,y)$. The first term attracts a generated sample $z$ toward the data, and the second repels it from other generated samples. Training regresses each output onto a stop-gradient target:
\begin{equation}
\calL_E(\theta)=\E_{\epsilon\sim\mu_0}\norm{f_\theta(\epsilon)-\StopGrad\!\left(f_\theta(\epsilon)+V_E(f_\theta(\epsilon))\right)}_2^2.
\label{eq:drifting-loss}
\end{equation}
Exchanging $p$ and $q$ negates the field, so $V_E=0$ when the model matches the data. The finite-batch estimator follows \citet{deng2026drifting} and is given in \appendixref{app:protocol}.

\paragraph{Topology in the loss.} For $\alpha,\tau\geq0$, define
\begin{equation}
G=I+\alpha L,\qquad \norm{u}_G^2=u^\top Gu,\qquad d_G(x,y)=\norm{x-y}_G,\qquad S=\exp(-\tau L).
\label{eq:hodge-geometry}
\end{equation}
Since $\norm{u}_G^2=\norm{u}_2^2+\alpha u^\top Lu$, this Hodge--Sobolev norm penalizes high-frequency components, and the heat kernel $S$ damps them. The Hodge loss uses topology in three places: $d_G$ sets the affinities $a^G_{p,q}$ and $b^G_{p,q}$, $S$ smooths each displacement, and $G$ measures the regression error:
\begin{equation}
\begin{aligned}
V_H(z)&=\E_{y\sim p}\!\left[a^G_{p,q}(z,y)S(y-z)\right]-\E_{y\sim q}\!\left[b^G_{p,q}(z,y)S(y-z)\right],\\
\calL_H(\theta)&=\E_{\epsilon\sim\mu_0}\norm{f_\theta(\epsilon)-\StopGrad\!\left(f_\theta(\epsilon)+V_H(f_\theta(\epsilon))\right)}_G^2.
\end{aligned}
\label{eq:hodge-loss}
\end{equation}
Unless stated otherwise, $\alpha=0.1$ and $\tau=0.5$. Proposition~\ref{prop:preconditioner} isolates the direct effect of $S$ and $G$ by holding the affinities and normalizers fixed at a single query.

\begin{proposition}[Fixed-affinity spectral response]
\label{prop:preconditioner}
Let $L=L^\top\succeq0$, $\alpha\geq0$, and $0\leq\tau<\infty$. Fix $z\in\R^d$, and suppose the Euclidean and Hodge fields at $z$ use the same coefficients $a(z,\cdot)$ and $b(z,\cdot)$ and the same scalar normalizers, and that both fields are finite. Let $\ell_E(w)=\norm{w-z-V_E(z)}_2^2$ and $\ell_H(w)=\norm{w-z-V_H(z)}_G^2$ be the per-query losses with their targets held fixed. Write $V_E(z)=\sum_{k=1}^d\beta_k u_k$ in an orthonormal eigenbasis with $Lu_k=\lambda_k u_k$. Then $V_H(z)=SV_E(z)$, and the descent directions at $w=z$ are
\begin{equation}
-\nabla\ell_E(z)=2\sum_{k=1}^d \beta_k u_k,\qquad
-\nabla\ell_H(z)=2\sum_{k=1}^d h(\lambda_k)\beta_k u_k,\qquad
h(\lambda)=(1+\alpha\lambda)e^{-\tau\lambda}.
\label{eq:response}
\end{equation}
Since $h>0$, the two directions have the same spectral support and vanish together.
\end{proposition}

The Hodge loss thus acts as a positive spectral preconditioner that keeps harmonic modes ($h(0)=1$) and damps high frequencies.

\paragraph{Topology in the predictor.} Independently of the loss, we can place $L$ in the predictor, $f_\theta(x)=r_\theta(x)+\gamma g_\theta(x;L)$, where $r_\theta$ is a residual MLP, $g_\theta$ is a shallow GCN \citep{kipf2017semi} whose layers map features $F\mapsto FW_0+LFW_1+b$ with a zero-initialized final layer, and $\gamma=0.1$. We compare four predictors. \emph{Graph-off} sets $\gamma=0$. \emph{Pointwise} keeps the trainable GCN branch but removes the propagation term $LFW_1$. \emph{Permuted} replaces $L$ by $PLP^\top$ for a random permutation $P$. \emph{Native} uses $L$. The permuted operator has the same spectrum, sparsity, and parameter count as $L$ but couples coordinates that are not neighbors in $K$, so it tests whether the predictor benefits from the domain itself.

\begin{figure}[t]
\centering
\begin{minipage}[c]{0.53\linewidth}\centering
\subfigure[\textbf{Hodge loss across $\alpha$ and $\tau$.}]{\includegraphics[width=\linewidth]{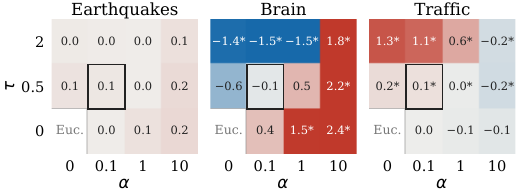}}
\end{minipage}\hfill
\begin{minipage}[c]{0.44\linewidth}\centering
\includegraphics[width=\linewidth]{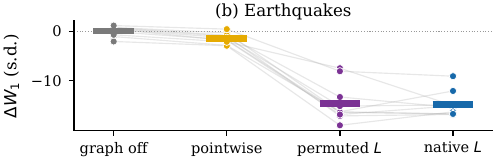}\\[2pt]
\includegraphics[width=\linewidth]{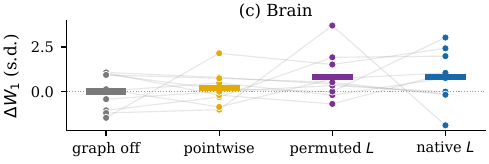}
\end{minipage}
\caption{\textbf{Neither the Hodge loss nor the domain's graph structure improves Drifting.} All panels show changes in $W_1$ in run-to-run s.d.; lower is better. (a) Hodge minus Euclidean loss for the MLP at each $(\alpha,\tau)$ (blue: Hodge better). The box marks the default, $\tau=0$ tests the metric alone, $\alpha=0$ tests heat smoothing alone, and $*$ denotes $p<0.05$. (b, c) GCN predictors relative to graph-off; native and permuted $L$ do not differ significantly on either task. Lines join seeds; bars are means.}
\label{fig:main-result}
\end{figure}

\section{Controlled experiments}
\label{sec:experiments}

\paragraph{Protocol.} We use the five TFM benchmarks of \citet{wyrwal2026topological}. Paired runs share data, initialization, and minibatches, and every configuration uses ten seeds (\appendixref{app:protocol}). As in these benchmarks, we score each model by the Wasserstein-1 distance $W_1$ between generated and real samples (lower is better), computed as a debiased Sinkhorn divergence with blur $0.05$ \citep{feydy2019sinkhorn}. Because scores differ in scale across tasks, we also report differences in \emph{run-to-run s.d.}, the standard deviation of a model's score across its ten training runs, so $1$ equals that noise.

\begin{table}[t]
\centering
\caption{\textbf{Endpoint $\SD$ for Euclidean and Hodge losses} ($\alpha=0.1$, $\tau=0.5$). Ten-seed mean (s.d.). The effect is Hodge minus Euclidean in run-to-run s.d.\ of the Euclidean runs, with its paired 95\% CI on the same scale.}
\label{tab:geometry}
\small
\setlength{\tabcolsep}{4pt}
\begin{tabular}{@{}llllrr@{}}
\toprule
Task & Predictor & Euclidean & Hodge & Effect [95\% CI] & $p$ \\
\midrule
Brain & MLP & $7.77\,(0.25)$ & $7.75\,(0.41)$ & $-0.08\;[-1.37,\,\phantom{-}1.21]$ & $0.89$ \\
Earthquakes & MLP & $5.741\,(0.002)$ & $5.741\,(0.002)$ & $0.06\;[-0.14,\,\phantom{-}0.26]$ & $0.51$ \\
Traffic & MLP & $1.269\,(0.004)$ & $1.269\,(0.004)$ & $0.12\;[\phantom{-}0.06,\,\phantom{-}0.19]$ & $0.002$ \\
Single cell & MLP & $0.0106\,(0.0004)$ & $0.0102\,(0.0005)$ & $-1.04\;[-2.17,\,\phantom{-}0.08]$ & $0.07$ \\
Ocean & MLP & $1.691\,(0.021)$ & $1.690\,(0.020)$ & $-0.02\;[-0.05,\,\phantom{-}0.01]$ & $0.15$ \\
\midrule
Brain & GCN & $6.06\,(0.12)$ & $6.06\,(0.10)$ & $0.01\;[-0.99,\,\phantom{-}1.00]$ & $0.99$ \\
Earthquakes & GCN & $5.721\,(0.003)$ & $5.722\,(0.005)$ & $0.28\;[-0.03,\,\phantom{-}0.59]$ & $0.07$ \\
\bottomrule
\end{tabular}
\end{table}

\paragraph{The Hodge loss does not help.} The Hodge--Euclidean difference is not significant on six of the seven comparisons in Table~\ref{tab:geometry}. On traffic the Hodge loss is $0.03\%$ worse ($p=0.002$), and on single cell it is $4\%$ better but not significantly so ($p=0.07$). On earthquakes, traffic, and ocean, any Hodge effect is smaller than run-to-run noise (95\% CI within $\pm0.6$ s.d.). Brain and single cell remain inconclusive.

\paragraph{Sweeping the Hodge loss.} Figure~\ref{fig:main-result}a varies $\alpha$ and $\tau$ for the MLP on three tasks, and no setting improves all three. No setting changes $\SD$ on earthquakes. Strong heat smoothing ($\tau=2$) lowers brain $\SD$ by about $4\%$ but raises traffic $\SD$ by about $0.3\%$, and a strong metric ($\alpha=10$) does the reverse, raising brain $\SD$ by up to $8\%$ and lowering traffic $\SD$ by $0.06\%$. The largest gain, on brain, does not carry over to the GCN, where $\tau=2$ raises $\SD$ by $1$--$3\%$ (\appendixref{app:sweep}).

\paragraph{Permuting the Laplacian.} On earthquakes (Figure~\ref{fig:main-result}b), the pointwise branch lowers $W_1$ by $1.5$ run-to-run s.d.\ relative to graph-off, and permuted propagation lowers it by a further $13$ s.d.\ (both $p<10^{-5}$). Native and permuted propagation do not differ ($-0.2$ s.d., 95\% CI $[-2.5,2.0]$, $p=0.83$). Since the permuted operator keeps the spectrum and sparsity of $L$, we attribute the gain to sparse mixing across coordinates. On brain (Figure~\ref{fig:main-result}c), which has a held-out test set, no GCN variant differs significantly from graph-off, and native and permuted propagation again agree ($0.0$ s.d., $p=0.98$; \appendixref{app:sweep}).

\paragraph{Comparison with I-TFM.} On brain data, we train I-TFM (TFM with independent coupling) using the authors' code and configuration. Euclidean Drifting with the GCN predictor reaches $\SD=6.06$ (s.d.\ $0.12$), against $6.41$ (s.d.\ $0.16$) for I-TFM (difference $-0.36$, Welch 95\% CI $[-0.49,-0.23]$), and is lower for $99$ of $100$ seed pairs. The same model without its graph branch reaches $5.99$ (Figure~\ref{fig:main-result}c), so the improvement does not depend on the graph. The $20$-epoch spectral MLP ($7.77$) does not beat I-TFM.

\section{Discussion}
\label{sec:discussion}

A TFM-style term $-\kappa LX_t$ cancels in a Drifting field (\appendixref{app:theory}), so topology can only reshape the discrepancies, which Proposition~\ref{prop:preconditioner} shows is spectral preconditioning. Euclidean Drifting is a strong, inexpensive baseline that topological generative models should report.

\paragraph{Limitations.} We test only degree-$0$ and degree-$1$ cochains over fixed domains, and no other topological regularizers. The sweep covers the MLP on three tasks, and because the metric and regression norm share $G$, it does not separate them. A synthetic target with prescribed Hodge components would be a more direct test, and the predictor controls cover only two tasks.

\clearpage
\bibliography{references}

\appendix

\section{Experimental protocol}
\label{app:protocol}

\begin{table}[ht]
\centering
\caption{Task definitions. $\dim K$ is the dimension of the domain used to build the operator. ``Ambient'' counts signal coordinates, and ``Active'' counts the coordinates retained for training. Evaluation reconstructs ambient signals, and single cell uses the PHATE clouds induced by mass rankings. Domain constructions follow \citet{yang2025topological}.}
\label{tab:tasks}
\small
\begin{tabular}{@{}lcclrr@{}}
\toprule
Task & $\dim K$ & Degree / operator & Distribution map & Ambient & Active \\
\midrule
Brain & $1$ & $0$ / $L_0$ & empirical $\to$ empirical & 360 & 360 \\
Earthquakes & $1$ & $0$ / $L_0$ & analytic $\to$ empirical & 576 & 576 \\
Traffic & $2$ & $1$ / $L_1$ & analytic $\to$ empirical & 340 & 340 \\
Single cell & $1$ & $0$ / $L_0$ & empirical $\to$ empirical & 18,203 & 256 \\
Ocean & $2$ & $1$ / $L_1$ & analytic $\to$ analytic & 19,023 & 500 \\
\bottomrule
\end{tabular}
\end{table}

The earthquake domain is a geodesic neighbor graph on vertices selected from a triangulated surface. The single-cell domain is a nearest-neighbor graph. Traffic and ocean use two-dimensional complexes, so their $L_1$ operators include contributions from both vertices and triangles. An edge experiment can therefore depend on two-dimensional incidence structure even though its signals have degree $1$. All domains are fixed throughout training and evaluation.

\subsection{Shared training and evaluation}

Each Hodge--Euclidean pair uses the same data split, active coordinates, initialization, optimizer, minibatches, checkpoint selection rule, and evaluation seed. The default predictor is a residual MLP with SwiGLU activations, trained for $20$ epochs with AdamW at learning rate $5\times10^{-4}$ and weight decay $10^{-4}$. The drift uses temperatures $\{0.02,0.05,0.2\}$. We normalize distances, normalize the affinities over rows and columns, add a large penalty to the distance between each generated sample and itself, and divide each temperature field by its RMS before summing. For the Hodge geometry, $\alpha=0.1$ and $\tau=0.5$.

We select the checkpoint with the smallest validation $\SD$. The exception is single cell, where the released protocol prescribes the debiased Sinkhorn-2 value at day $4$. Paired confidence intervals and $p$-values are two-sided $t$ procedures on the differences across the ten matched seeds. They are pointwise intervals, conditional on the fixed data and evaluation protocol. A non-significant difference leaves its sign unresolved. To convey its size, Table~\ref{tab:geometry} and Figure~\ref{fig:main-result}a divide each paired difference and its interval by the run-to-run s.d., the sample s.d.\ of the ten Euclidean runs. This is a descriptive scale and not a formal equivalence test.

\subsection{Brain fMRI}

The brain experiment transports between empirical laws on $360$ brain regions, using $1{,}190$ paired signals split into $857$ training, $95$ validation, and $238$ test examples. The MLP and GCN have width $512$ and depth $4$. The MLP uses batches of $32$ and gradient clipping at norm $1$. The GCN is trained for $100$ epochs and has $2.13$ million parameters. The I-TFM baseline uses independent coupling, the topological heat reference with $\kappa=2$ and noise level $\sigma=0$, and the released training configuration. It is trained and evaluated on the same split with ten seeds.

\subsection{Earthquakes}

The earthquake experiment transports a coordinatewise Gaussian law to the empirical law of $28$ yearly fields in $\R^{576}$. Following the released benchmark, the Gaussian is fitted to these fields, and the same fields are reused for training, validation, and testing. The MLP and GCN have width $256$ and depth $3$ and are trained for $20$ epochs with batches of $28$. The graph-off, pointwise, and propagating predictors have $706{,}624$, $707{,}393$, and $707{,}905$ active parameters, out of $707{,}905$ instantiated parameters in every case. The permuted operator for seed $s$ is $P_sLP_s^\top$, with permutation seed $10{,}000+s$. To confirm that propagation remains active after training, we measure $\E\norm{LFW_1}_2/\E\norm{FW_0+b}_2$ in the two GCN layers over $28$ source draws for each of ten seeds. The mean ratios are $(0.269,1.32)$ for native and $(0.269,1.44)$ for permuted propagation. Because all $28$ fields are used to fit the source law, select checkpoints, and evaluate endpoints, the predictor control measures in-sample distribution matching.

\subsection{Traffic}

The traffic experiment transports a coordinatewise Gaussian fitted to the training targets toward an empirical law of edge flows in $\R^{340}$. The $16{,}992$ observations are split into $12{,}235$ training, $1{,}359$ validation, and $3{,}398$ test examples. Before eigendecomposition, we express the normalized Hodge operator in coordinates where it is self-adjoint. The MLP has width $256$ and depth $3$, uses batches of $512$, and is evaluated on the complete held-out target.

\subsection{Single cell}

The single-cell experiment transports one mass vector to another over $18{,}203$ cells, with both represented in the $256$ released eigenmodes. A generated example is therefore one predicted mass vector, not an independent cell. Its mass ranking assigns predicted day labels to cells. For each day, $\SD$ compares the resulting PHATE cloud with the true cloud for that day, and Table~\ref{tab:geometry} reports day $4$. The MLP has width $256$ and depth $3$ and an output layer initialized at scale $0.03$. It uses batches of one. The only negative sample is then the query itself, whose displacement is zero, so the repelling term vanishes and the field reduces to attraction toward the data.

\subsection{Ocean}

The ocean experiment transports between analytic Gaussian laws on flows over a complex with $19{,}023$ edges, represented by the released $94$ gradient modes and $406$ curl modes. The MLP has width $256$ and depth $3$ and uses batches of $512$. Each seed draws $25{,}600$ target samples for training and independent samples for validation and testing. Endpoint evaluation uses fresh sets of $512$ source and $512$ target samples.

\subsection{Hyperparameter sweep}
\label{app:sweep}

The sweep reruns the MLP Hodge variant on earthquakes, brain, and traffic at every $(\alpha,\tau)\in\{0,0.1,1,10\}\times\{0,0.5,2\}$ except $(0,0)$, which equals the Euclidean loss, and the default $(0.1,0.5)$, which is reported in Table~\ref{tab:geometry}. Every other setting matches the main experiments, and all runs use the code revision of the main experiments from a clean checkout. Each cell is paired by seed with the Euclidean runs of Table~\ref{tab:geometry}. At the default $\alpha=0.1$ the metric barely changes the affinities (\appendixref{app:diagnostics}), which the sweep extends to $\alpha=10$.

Rerunning seed $0$ reproduces the earthquake and traffic records to within $10^{-6}$. Brain training is not bitwise deterministic on the GPU: an identical rerun of seed $0$ changed $\SD$ from $8.24$ to $7.53$, so pairing removes less variation on brain. We therefore also compared brain against ten fresh Euclidean runs (mean $7.68$). Against them, $\tau=2$ still lowers $\SD$ by $3.5$--$3.8\%$ (Welch $p\leq0.03$), and $\alpha=10,\tau=0$ raises it by $8.7\%$. The brain predictor controls use the brain GCN configuration with the same graph-off, pointwise, and permuted modifications as on earthquakes. Relative to graph-off (mean $W_1=5.99$, run-to-run s.d.\ $0.08$), pointwise, permuted, and native propagation reach $6.01$, $6.06$, and $6.06$, all within noise ($p\geq0.18$). For the brain GCN, $\tau=2$ gives $\SD=6.12$ with $\alpha=0.1$ and $6.18$ with $\alpha=0$, against $6.06$ for the Euclidean runs of Table~\ref{tab:geometry} and $6.00$ for ten fresh Euclidean runs.

\subsection{Finite-batch field}

Let $z_1,\dots,z_n$ be generated queries, and set $y_j^+=y_j$ (data) and $y_j^-=z_j$ (generated samples). For geometry $J\in\{E,H\}$, let $d_J$ and $D_J$ denote the distance and the displacement, so $D_E(z,y)=y-z$ and $D_H(z,y)=S(y-z)$. Distances are floored at $10^{-4}$, and $\bar d_J$ is their mean over all $2n^2$ entries. We set
\begin{equation}
\delta_{ij}^{+,J}=\frac{d_J(z_i,y_j^+)}{\max\{\bar d_J,10^{-3}\}},\qquad \delta_{ij}^{-,J}=\frac{d_J(z_i,y_j^-)}{\max\{\bar d_J,10^{-3}\}}+100\cdot\mathbf{1}_{i=j}.
\label{eq:finite-distances}
\end{equation}
For temperature $T$ and branch $s\in\{+,-\}$, the row and column normalizations over the joint positive and negative matrix are
\begin{equation}
R_{ij}^{s,T}=\frac{e^{-\delta_{ij}^{s,J}/T}}{\sum_{r\in\{+,-\}}\sum_k e^{-\delta_{ik}^{r,J}/T}},\qquad
C_{ij}^{s,T}=\frac{e^{-\delta_{ij}^{s,J}/T}}{\sum_k e^{-\delta_{kj}^{s,J}/T}},
\label{eq:finite-affinities}
\end{equation}
and the affinities are $A_{ij}^{s,T}=\sqrt{\max\{R_{ij}^{s,T}C_{ij}^{s,T},10^{-6}\}}$. With row masses $\omega_i^{+,T}=\sum_jA_{ij}^{+,T}$ and $\omega_i^{-,T}=\sum_jA_{ij}^{-,T}$, the unnormalized field is
\begin{equation}
U_{i,J}^T=\omega_i^{-,T}\sum_jA_{ij}^{+,T}D_J(z_i,y_j^+)-\omega_i^{+,T}\sum_jA_{ij}^{-,T}D_J(z_i,y_j^-).
\label{eq:finite-field}
\end{equation}
With the scale $s_J=\max\{\bar d_J/\sqrt d,10^{-3}\}$, each temperature field is divided by its RMS and the results are summed:
\begin{equation*}
\sigma_{J,T}^2=\frac{1}{nd}\sum_i\left\|\frac{U_{i,J}^T}{s_J}\right\|_J^2,\qquad V_{i,J}=\sum_T\frac{U_{i,J}^T/s_J}{\sqrt{\max\{\sigma_{J,T}^2,10^{-8}\}}}.
\end{equation*}
Regression is performed in the scaled coordinates $z_i/s_J$, with per-query loss
\begin{equation*}
\frac{1}{d}\left\|\frac{z_i}{s_J}-\StopGrad\!\left(\frac{z_i}{s_J}+V_{i,J}\right)\right\|_J^2,
\end{equation*}
where $s_J$ is treated as constant under differentiation. Padded batches use the same equations after masking invalid rows and columns, with $n$ replaced by the number of valid queries.

\subsection{Inference timing}
\label{app:timing}

Timing uses batches of $512$ on the GPU listed in \appendixref{app:reproducibility}, with one warm-up and five repetitions per seed, after compilation. Across ten seeds, the mean per-seed median is $4.4$~ms for the one-evaluation Drifting generator and $1{,}408$~ms for I-TFM. The mean per-seed ratio is $325\times$ (s.d.\ $49$). I-TFM integrates its SDE with Euler--Maruyama on $1{,}000$ uniform steps from $t_0=10^{-4}$. These are end-to-end timings of two different implementations. Drifting runs in single precision in JAX, while I-TFM runs in PyTorch with its released double-precision configuration, which accounts for part of the gap.

\section{Strength of the interventions}
\label{app:diagnostics}

Table~\ref{tab:intervention} evaluates the $\alpha=0.1$, $\tau=0.5$ configuration at the matched seed-$0$ initialization, on the first training batch of each task in its active spectral coordinates. Correlations exclude the penalized self-distances. Affinities and fields follow \equationref{eq:finite-affinities} and \equationref{eq:finite-field}. The single-cell batch has one query, so its distance and affinity correlations are undefined.

\begin{table}[ht]
\centering
\caption{Strength of the Hodge intervention at $\alpha=0.1$ and $\tau=0.5$. The second column gives $h$ at the smallest, median, and largest eigenvalue. $r_d$ and $r_A$ are the correlations between Euclidean and Hodge distances and affinities, $c_V$ is the mean cosine between the two fields, and $\rho_V=\|V_H-V_E\|_2/\|V_E\|_2$.}
\label{tab:intervention}
\scriptsize
\begin{tabular}{@{}lccccc@{}}
\toprule
Task & $h(\lambda_{\min},\lambda_{\mathrm{med}},\lambda_{\max})$ & $r_d$ & $r_A$ & $c_V$ & $\rho_V$ \\
\midrule
Brain & $(1.000,0.658,0.543)$ & $0.99996$ & $0.99994$ & $0.995$ & $0.106$ \\
Earthquakes & $(1.000,0.641,0.441)$ & $1.00000$ & $1.00000$ & $0.968$ & $0.250$ \\
Traffic & $(1.000,0.810,0.536)$ & $1.00000$ & $1.00000$ & $0.985$ & $0.172$ \\
Single cell & $(1.000,0.584,0.483)$ & -- & -- & $0.959$ & $0.285$ \\
Ocean & $(0.999,0.859,0.813)$ & $1.00000$ & $0.99995$ & $0.997$ & $0.082$ \\
\bottomrule
\end{tabular}
\end{table}

\section{Proofs and cancellation details}
\label{app:theory}

\begin{proof}[Proof of Proposition~\ref{prop:preconditioner}]
Both expectations are finite by assumption. Since $S$ is linear on $\R^d$, it commutes with the expectations:
\begin{equation*}
V_H(z)=S\left(\E_{y\sim p}[a(z,y)(y-z)]-\E_{y\sim q}[b(z,y)(y-z)]\right)=SV_E(z).
\end{equation*}
Because $G$ is symmetric, $\nabla\ell_E(w)=2(w-z-V_E(z))$ and $\nabla\ell_H(w)=2G(w-z-V_H(z))$. At $w=z$ these give $-\nabla\ell_E(z)=2V_E(z)$ and $-\nabla\ell_H(z)=2GSV_E(z)$. By the spectral theorem, $L$ has an orthonormal eigenbasis with $\lambda_k\geq0$, and $GSu_k=(1+\alpha\lambda_k)\exp(-\tau\lambda_k)u_k=h(\lambda_k)u_k$. Expanding $V_E(z)=\sum_k\beta_ku_k$ gives \equationref{eq:response}. Finally, $h(\lambda_k)>0$ for every $k$, so $\beta_k=0$ exactly when $h(\lambda_k)\beta_k=0$.
\end{proof}

\paragraph{Exchange symmetry.} For the population field in \equationref{eq:euclidean-field}, the identity $a_{p,q}=b_{q,p}$ gives
\begin{equation*}
V_{q,p}(z)=\E_{y\sim q}[b_{p,q}(z,y)(y-z)]-\E_{y\sim p}[a_{p,q}(z,y)(y-z)]=-V_{p,q}(z),
\end{equation*}
and setting $p=q$ gives a zero field. The same argument applies to the Hodge field, because the same map $S$ acts on both displacements. A scalar normalization preserves this identity if its value does not change when the laws are exchanged. The finite-batch estimator does not have this exact symmetry. Its queries are generated samples, its column normalization uses the query cohort, and the self-distance penalty applies only to the negative branch. The population cancellation therefore holds for the estimator only approximately.

\paragraph{Retained subspaces.} For finite $\tau$, every eigenvalue of $S$ is $\exp(-\tau\lambda)>0$, so heat smoothing is invertible on the active coordinates. For single cell and ocean, these coordinates span a proper subspace, because only the released modes are modeled. Proposition~\ref{prop:preconditioner} applies within that subspace.

\paragraph{Cancellation of query-only drifts.} Suppose we add the same vector $\xi(z_i)$ to every displacement of query $z_i$ in \equationref{eq:finite-field}. The net addition is $(\omega_i^{-,T}\omega_i^{+,T}-\omega_i^{+,T}\omega_i^{-,T})\xi(z_i)=0$. A path method can add $-\kappa LX_t$ to its dynamics directly, but the balanced displacement construction cancels any such query-only term.

\section{Software and hardware}
\label{app:reproducibility}

\subsection{Software packages}
\label{app:software}

Native models are implemented and trained in Python~3.13.9 with \href{https://github.com/jax-ml/jax}{JAX}~0.10.0~\citep{jax2018github}, \href{https://github.com/patrick-kidger/equinox}{Equinox}~0.13.8~\citep{kidger2021equinox}, and \href{https://github.com/google-deepmind/optax}{Optax}~0.2.8~\citep{deepmind2020jax}. Data are loaded with \href{https://github.com/smorad/cyreal}{Cyreal}~0.2.1. Endpoint metrics are computed with \href{https://github.com/jeanfeydy/geomloss}{GeomLoss}~0.3.1~\citep{feydy2019sinkhorn} on \href{https://github.com/pytorch/pytorch}{PyTorch}~2.12.1~\citep{paszke2019pytorch}.

\subsection{Hardware}

Experiments ran on an AMD Ryzen 9950X3D CPU with 64~GB of 4000~MHz RAM and an NVIDIA RTX 5080 GPU with 16~GB of memory.

\end{document}